\documentclass[10pt,twocolumn,letterpaper]{article}

\usepackage{cvpr}
\usepackage{times}
\usepackage{epsfig}
\usepackage{graphicx}
\usepackage{amsmath}
\usepackage{amssymb}
\usepackage{xcolor}
\usepackage{booktabs}
\usepackage{paralist}
\usepackage{bbm}
\usepackage{multirow}
\usepackage{xspace}
\usepackage{enumitem}
\usepackage{bbm} 
\usepackage{siunitx}
\usepackage{textcomp}
\usepackage{adjustbox}
\usepackage{cellspace} 
\usepackage[pagebackref=true,breaklinks=true,letterpaper=true,colorlinks,bookmarks=false]{hyperref}
 \cvprfinalcopy 

\def\cvprPaperID{*} 

\makeatletter
\DeclareRobustCommand\onedot{\futurelet\@let@token\@onedot}
\def\@onedot{\ifx\@let@token.\else.\null\fi\xspace}
\def\eg{\emph{e.g}\onedot} 
\def\ie{\emph{i.e}\onedot} 
\def\cf{\emph{c.f}\onedot} 
 \def\vs{\emph{vs}\onedot} 
\def\wrt{\emph{w.r.t}\onedot} 

\makeatother

\def\be{\begin{equation}}
\def\ee{\end{equation}}
\def\bea{\begin{eqnarray}}
\def\eea{\end{eqnarray}}
\def\fig#1{Fig.~\ref{fig:#1}}
\def\tab#1{Table~\ref{tab:#1}}
\def\sect#1{Sec.~\ref{sec:#1}}

\def\mypar#1{\vspace{1mm}{\noindent\bf #1.}\hspace{1mm}}

\ifcvprfinal\fi
\begin{document}
\title{Hierarchical Scene Coordinate Classification and Regression\\
for Visual  Localization}


\author{Xiaotian Li$^1$ \quad
Shuzhe Wang$^1$ \quad
Yi Zhao$^1$ \quad
Jakob Verbeek$^2$\thanks{Work done while JV was at INRIA.} \quad
Juho Kannala$^1$ \\
$^1$Aalto University \quad 
$^2$Facebook AI Research
}

\maketitle
\thispagestyle{empty}

\begin{abstract}
Visual localization is critical to many applications in computer vision and robotics. To address single-image RGB localization, state-of-the-art feature-based methods match local descriptors between a query image and a pre-built 3D model. Recently, deep neural networks have been exploited to regress the mapping between raw pixels and 3D coordinates in the scene, and thus the matching is implicitly performed by the forward pass through the network. However, in a large and ambiguous environment, learning such a regression task directly can be difficult for a single network. In this work, we present a new hierarchical scene coordinate network to predict pixel scene coordinates in a coarse-to-fine manner from a single RGB image. The  network consists of  a series of output layers, each of them conditioned on the previous ones. The final output layer predicts the 3D coordinates and the others produce progressively finer discrete location labels. The proposed method outperforms the baseline regression-only network and allows us to train  compact models which scale robustly to large environments. It sets a new state-of-the-art for single-image RGB localization performance on  the 7-Scenes, 12-Scenes, Cambridge Landmarks  datasets, and three combined scenes. 
Moreover, for large-scale outdoor localization on the Aachen Day-Night dataset, we present a hybrid approach which outperforms existing scene coordinate regression methods, and reduces significantly the performance gap w.r.t.\ explicit feature matching methods.\footnote {Code and materials available at \url{https://aaltovision.github.io/hscnet}.
}
\end{abstract}

%

\section{Introduction}
Visual localization aims at estimating precise six degree-of-freedom (6-DoF) camera pose with respect to a known environment. 
It is a fundamental component of  many intelligent autonomous systems and applications in computer vision and robotics, \eg, augmented reality, autonomous driving, or camera-based indoor localization for personal assistants.
Commonly used visual localization methods rely on matching local visual descriptors~\cite{SattlerLK11,SattlerLK17}. Correspondences are typically established between 2D interest points in the query and 3D points in the pre-built structure-from-motion model~\cite{schoenberger2016sfm,schoenberger2016mvs} with nearest neighbor search, and the 6-DoF camera pose of the query can then be computed from the correspondences. 

Instead of explicitly establishing 2D-3D correspondences via matching descriptors, scene coordinate regression methods directly regress 3D scene coordinates from an image~\cite{Brachmann_2017_CVPR,Brachmann_2018_CVPR,brachmann2020visual,SCoRF}. 
In this way, correspondences between 2D points in the image and 3D points in the scene can be obtained densely without feature detection and description, and explicit matching. In addition, no  descriptor database is required at test time since the model weights encode the scene representation implicitly. It was experimentally  shown that recent CNN-based scene coordinate regression methods achieve better localization performance on small-scale  datasets compared to the state-of-the-art feature-based methods~\cite{Brachmann_2018_CVPR}. The high accuracy and the compact representation of a dense scene model make scene coordinate regression approach an interesting alternative to the classic feature-based approach.

\begin{figure*}
\begin{center}
\includegraphics[width=0.75\linewidth]{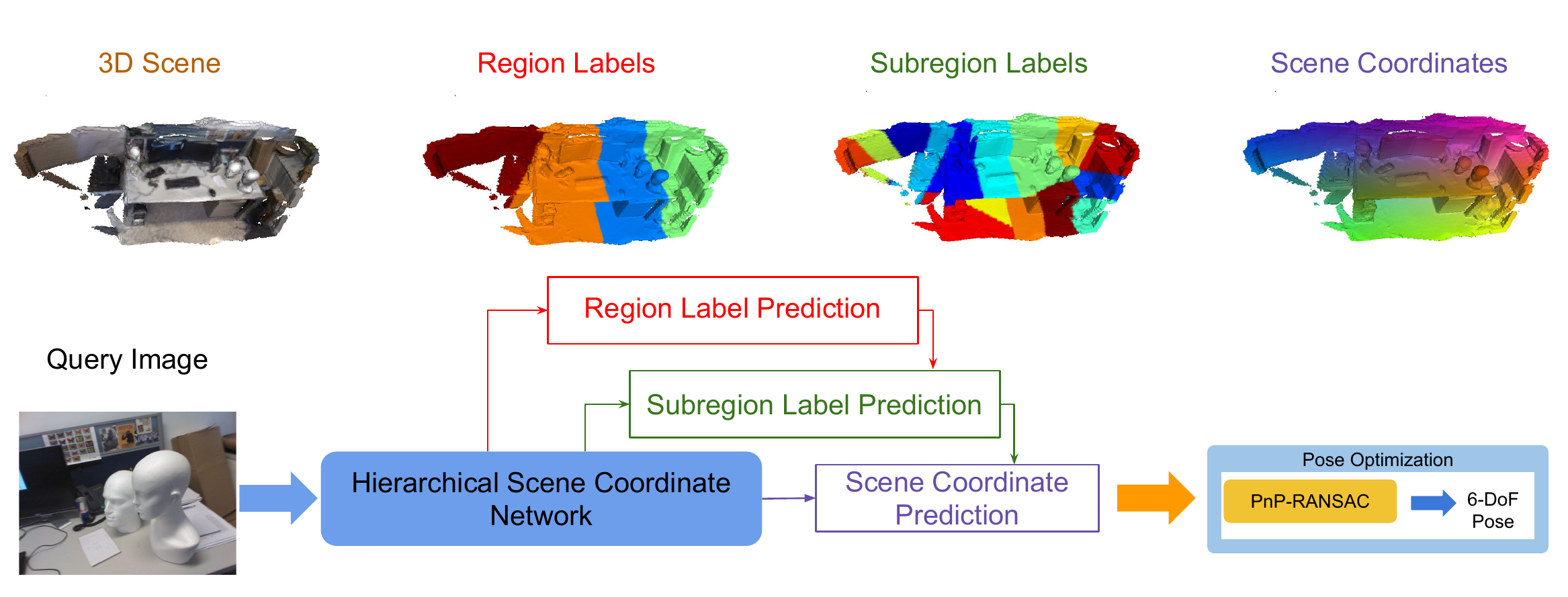}
\end{center}
\caption{Overview of our single-image RGB localization approach based on hierarchical scene coordinate prediction, here using 3 levels.}
\label{fig:teaser}
\end{figure*}	

However, most existing scene coordinate regression methods can only be adopted on small-scale scenes. Typically, scene coordinate regression networks are designed to  have a limited receptive field~\cite{Brachmann_2017_CVPR,Brachmann_2018_CVPR}, \ie only a small local image patch is considered for each scene coordinate prediction. 
This allows the network to generalize well from   limited training data, since local patch appearance is more stable \wrt viewpoint change. 
On the other hand, a limited receptive field size can lead to ambiguous patterns in the scene, especially in large-scale environments, caused by visual similarity between local image patches.
 Due to these ambiguities, it is harder for the network to accurately model the regression problem,  resulting in inferior performance at test time.
Using larger receptive field sizes, up to the full image, to regress the coordinates can mitigate the issues caused by ambiguities. 
This, however, has  been shown to be prone to overfitting the larger input patterns in the case of limited training data, even if data augmentation  alleviates this problem to some extent~\cite{Li2018}.

In contrast, in this work, we overcome the   ambiguities due to  small receptive fields  by conditioning on  discrete location labels around each pixel. 
During training, the labels are obtained by a coarse quantization of the ground-truth 3D coordinates.
At test time, the  location labels for each pixel are  obtained using dense classification networks, which can more easily deal with the location ambiguity since they are trained using the cross-entropy classification loss which permits a multi-modal prediction in 3D space.  
Our model allows for several classification layers, using progressively finer location labels, obtained through hierarchical clustering of the ground-truth 3D point cloud data. 
Our hierarchical coarse-to-fine architecture is implemented using  conditioning layers that are related to the FiLM  architecture~\cite{FILM},  resulting in a compact model.    
See \fig{teaser} for a schematic overview of our approach.

We validate our approach by comparing it to a regression-only network, which lacks the hierarchical coarse-to-fine structure.
We present results on three datasets used in previous works: %
7-Scenes~\cite{SCoRF}, 12-Scenes~\cite{valentin2016learning}, and Cambridge Landmarks~\cite{kendall2015convolutional}. 
Our approach shows consistently better performance and achieves state-of-the-art results %
for single-image RGB localization.
Moreover, by compiling the 7-Scenes and 12-Scenes datasets into single large scenes, and using the Aachen Day-Night dataset~\cite{sattler2018benchmarking,sattler2012image}, we show that our approach scales more robustly to larger environments. 

In summary, our contributions  are as follows:
\begin{compactitem}
  \item We introduce a new 
  hierarchical coarse-to-fine conditioning 
  architecture for scene coordinate prediction, 
  which improves the performance and scalability over   a baseline regression-only network.
  
  \item We show that our novel approach achieves state-of-the-art results for single-image RGB localization on three benchmark datasets and it allows us to train single compact models which scale robustly to large scenes.
  \item For large-scale outdoor localization, we present a hybrid approach built upon our network, which reduces significantly the gap to feature-based methods.
\end{compactitem}

\section{Related Work}
\mypar{Visual localization}
Visual localization aims at predicting 6-DoF camera pose for a given query image. 
To obtain precise 6-DoF camera pose, visual localization methods are typically structure-based, \ie they rely on 2D-3D correspondences between 2D image positions and 3D scene coordinates.  
With the established 2D-3D correspondences, a RANSAC~\cite{RANSAC} optimization scheme is responsible for producing the final pose estimation. 
The correspondences are typically obtained by matching local features such as SIFT~\cite{SIFT}, and many matching and filtering techniques have been proposed,  which enable efficient and robust city-scale localization~\cite{Cheng_2019_ICCV,Larsson_2019_ICCV,Middelberg2014,SattlerLK17,Svarm2017,Toft_2018_ECCV}.

Image retrieval can also be used for visual localization~\cite{NetVLAD}. The pose of the query image can be directly approximated by the most similar retrieved database image. Since compact image-level descriptors are used for matching, image retrieval methods can scale to very large environments. The retrieval methods can be combined with structure-based methods~\cite{sarlin2019coarse,sarlin2018leveraging,Sattler17cvpr,inloc,Yang_2019_ICCV} or relative pose estimation~\cite{Balntas_2018_ECCV,Ding_2019_ICCV,LaskarMKK17} to predict precise poses. Typically, the retrieval step helps restrict the search space, leading to faster and more accurate localization.

In recent years, learning-based localization approaches have been explored. One popular direction is to replace the entire localization pipeline with a single neural network. PoseNet~\cite{kendall2015convolutional} and its variants~\cite{mapnet2018,KendallC15bay,Kendall_2017_CVPR,MelekhovYKR17,Walch_2017_ICCV} directly regress the camera pose from a query image. Recently, however, it was demonstrated that direct  pose regression yields results more similar to pose approximation via image retrieval than to accurate pose estimation via 3D structure~\cite{Sattler2019}. 
Therefore, these methods are still outperformed by structure-based methods. By fusing estimated pose information from the previous frame, \cite{radwan2018vlocnet++,valada2018deep} achieve  better performance, but require sequences of images rather than single images.

\mypar{Scene coordinate regression}
Instead of learning the entire pipeline, scene coordinate regression methods learn the first stage of the pipeline in the structure-based approaches. Namely, either a random forest~\cite{BrachmannMKYGR16,cavallari2019real,CavallariGLVST17,Guzman-RiveraKGSSFI14,Massiceti2017,meng2017backtracking,meng2018exploiting,SCoRF,ValentinNSFIT15} or a neural network~\cite{Brachmann_2017_CVPR,Brachmann_2018_CVPR,Brachmann_2019_ICCV,Brachmann_2019_ICCV_NG,brachmann2020visual,Budvytis2019,bui2018scene,Cavallari_corr_19,Li2018,Li_Ylioinas_Verbeek_Kannala_2018,Massiceti2017} is trained to directly predict 3D scene coordinates for the pixels and thus the 2D-3D correspondences are established. These methods do not explicitly rely on feature detection, description and matching, and are able to provide correspondences densely. 
They are more accurate than traditional feature-based methods at small and medium scale, but usually do not scale well to larger scenes~\cite{Brachmann_2018_CVPR,Brachmann_2019_ICCV}. 
In order to generalize well to novel viewpoints, these methods typically rely on only local image patches to produce the scene coordinate predictions. 
However, this may introduce ambiguities due to similar local appearances, especially when the scale of the scene is large. 
To resolve local appearance ambiguities,  we introduce element-wise conditioning layers to modulate the intermediate feature maps of the network using coarse discrete location information.
We show this leads to better localization performance, and we can robustly scale to larger environments.

\mypar{Joint classification-regression}
Joint classification-regression frameworks have been proved effective in solving various vision tasks. 
For example, \cite{rogez17cvpr,rogez19pami} proposed a classification-regression approach for human pose estimation from single images. %
In~\cite{BrachmannMKYGR16},  a joint classification-regression forest is trained to predict scene identifiers and scene coordinates. In~\cite{Weinzaepfel_2019_CVPR}, a CNN is used to detect and segment a predefined set of planar Objects-of-Interest (OOIs), and then, to regress dense matches to their reference images. %
In~\cite{Budvytis2019}, scene coordinate regression is formulated as two separate tasks of object instance recognition and local coordinate regression. In~\cite{Brachmann_2019_ICCV}, multiple scene coordinate regression networks are trained as a mixture of experts along with a  gating network which assesses the relevance of each expert for a given input, and the final pose estimate is obtained using a novel RANSAC framework, \ie, Expert Sample Consensus
(ESAC). 
In contrast to existing approaches, in our work, we use spatially dense discrete location labels   defined for all pixels, and propose FiLM-like~\cite{FILM} conditioning layers to propagate information in the hierarchy. 
We show that our novel framework allows us to achieve high localization accuracy with one single compact model.

\section{Hierarchical Scene Coordinate Prediction}
\label{sec:method}

We now describe our  coarse-to-fine hierarchical  scene coordinate prediction approach. 
Note that we address single-image RGB  localization, as in \eg~\cite{Brachmann_2018_CVPR,Brachmann_2019_ICCV, Brachmann_2019_ICCV_NG,Li_Ylioinas_Verbeek_Kannala_2018},  rather than using RGB-D images~\cite{Cavallari_corr_19,cavallari2019real,CavallariGLVST17,Guzman-RiveraKGSSFI14,meng2018exploiting,SCoRF,ValentinNSFIT15}, or image sequences~\cite{radwan2018vlocnet++,valada2018deep}.

\mypar{Hierarchical joint learning framework}
To define  hierarchical discrete location labels, we  hierarchically partition the ground-truth 3D point cloud data. This step can be done, \eg, with  k-means. 
In this way, in addition to the ground-truth 3D scene coordinates, each  pixel in a training image is also associated with a number of labels, from coarse to fine, obtained at different levels of the clustering hierarchy. Then, for each level, our network has a corresponding classification  layer which  for all pixels predicts the discrete location labels at that level. 
Besides the classification layers, we include a final regression layer to predict the continuous 3D scene coordinates for the  pixels,  generating  putative 2D-3D matches. 
To propagate the coarse location information to inform the predictions at finer levels, we introduce conditioning layers before each classification/regression layer. 
Note that we condition on the ground truth label maps during training, and condition on the predicted label maps at test time.

Since the predictions in each classification layer are conditioned on all preceding label maps, at each particular classification layer, it suffices to  predict  the label branch at that level. %
For example, for a three-level classification hierarchy, with branching factor $k$, we classify across only $k$ labels at each level. 
Similar to~\cite{Budvytis2019}, instead of directly regressing the absolute coordinates, we regress the relative positions to the cluster centers in 3D space at the finest level. %
This accelerates convergence of network training~\cite{Budvytis2019}. Note that this hierarchical scene coordinate learning framework also allows a classification-only variant. That is, if we have  fine enough location labels before the regression layer, we can simply use the cluster centers as the scene coordinates predictions without performing a  final regression step.

We  design the network to be global-to-local, which means that finer output layers have smaller receptive fields in the input image. 
This allows the network to use more global information at coarser levels, while conditioning on location labels to disambiguate the local appearances at finer levels. 
Note that at test time, the receptive fields of the finer output  layers  are also large, as they depend on the discrete location labels which are predicted from the input at test time, rather than fixed as during training.

\begin{figure*}
  \includegraphics[width=0.75\linewidth]{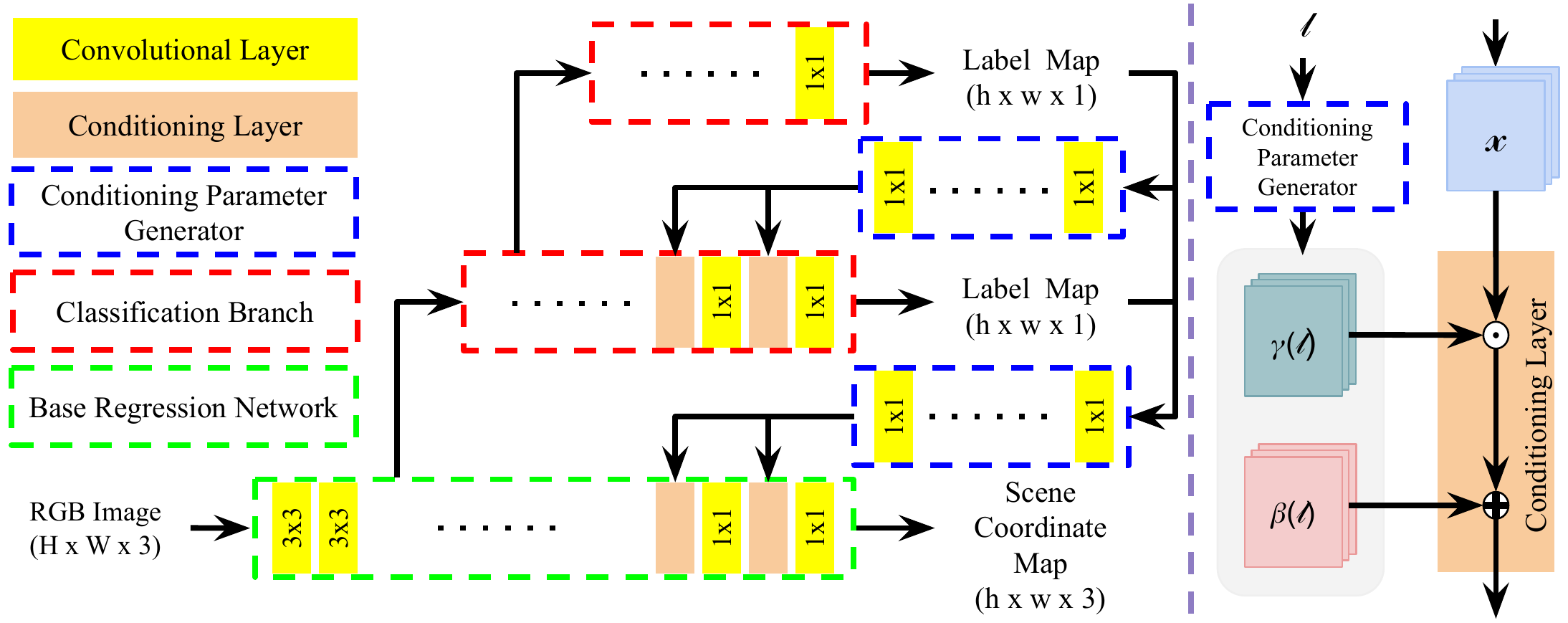}
   \centering
  \caption{Left:  Architecture of our hierarchical scene coordinate network (3-level).
Right: Architecture of the conditioning layer.  }
  \label{fig:archi}
\end{figure*}	

\mypar{Conditioning layers}
To make use of the discrete location label information predicted by the network at coarser levels, these predictions  should be fed back to the finer levels. 
Inspired by the Feature-wise Linear Modulation (FiLM) conditioning method~\cite{FILM}, we introduce conditioning layers just before each of the output layers. 
A conditioning parameter generator takes the predicted label map $\ell$  as input, outputs a set of scaling and shifting parameters $\gamma(\ell)$ and $\beta(\ell)$, and these parameters are fed into the conditioning layer to apply linear transformation to the input feature map. 
Unlike  FiLM layers, however, which perform the same channel-wise modulation across the entire feature map, our conditioning layers perform a linear modulation per spatial position, \ie, element-wise multiplication and addition as shown in \fig{archi} (right).
Therefore, instead of  vectors, the output parameters $\gamma(\ell)$ and $\beta(\ell)$ from a generator are feature maps of the same (height, width, channel) dimensions as the input feature map of the corresponding conditioning layer. 
More formally, given the input feature map $x$, the scaling and shifting parameters $\gamma(\ell)$ and $\beta(\ell)$, the linear modulation can be written as:
\begin{equation}
f(x,\ell) = 	  \gamma(\ell)\odot x + \beta(\ell),
\end{equation}
where $\odot$ denotes the Hadamard product.
In addition, the generators consist of only $1\!\times\!1$  convolutional layers so that each pixel is conditioned on its own location labels. 
We use an ELU non-linearity~\cite{ClevertUH15} after the feature modulation.  

\mypar{Network architecture}
In our main experiments we use 3-level hierarchy for all the datasets, \ie our  network has two classification output layers and one regression output layer. 
The overall architecture of this network is shown in \fig{archi} (left). 
The first classification branch predicts the coarse location labels, and the second one predicts  the fine labels.
We use strided convolution, upconvolution and dilated convolution for the two classification branches to enlarge the size of the receptive field, while preserving the output resolution. 
All the layers after the conditioning layers have kernel size of $1\!\times\!1$  such that the label conditioning is applied locally. %
More details on the architecture are provided in the supplementary material.

\mypar{Loss function}
Our network predicts location labels and regresses scene coordinates at the same time. 
Therefore, we need both a regression loss and a classification loss during training. 
For the regression task, we minimize the Euclidean distance between  predicted scene coordinates $\hat{y}$ and ground truth scene coordinates $y$,
\begin{equation}
\label{eucliloss}
\mathcal{L}_{r}  = \sum_i\| y_i-\hat{y}_i\|_{2},
\end{equation}
where $i$ ranges over the pixels in the image.
For the classification task, we use cross-entropy loss at each level, \ie
\begin{equation}
\label{celoss}
\mathcal{L}_{c}^j  = -\sum_i \left(\ell_i^j\right)^\top \log \hat{\ell}_i^j, 
\end{equation}
where $\ell_i^j$ denotes the one-hot coding of the ground-truth label of pixel $i$ at level $j$, and $\hat{\ell}_i^j$ denotes the vector of predicted label probabilities for the same pixel, and the logarithm is applied element-wise. In the case of 3-level hierarchy, the final loss function is given by
\begin{equation}
\label{loss}
\mathcal{L}  = w_1\mathcal{L}_{c}^1 + w_2\mathcal{L}_{c}^2 + w_3\mathcal{L}_{r},
\end{equation}
where $w_1$, $w_2$, $w_3$ are  weights for the loss terms. We found that the accuracy of the final regression prediction is crucial to localization performance, and thus a large value should be set for the regression loss. Details on the weights and training procedure are provided in the supplementary material. Note that, as mentioned before, our hierarchical joint learning framework also allows a classification-only variant, by using a finer label hierarchy. %

\section{Experimental Evaluation}
\label{sec:result}

In this section, we present our experimental setup and evaluation results  on standard visual localization datasets.

\begin{table*}[t!]
\renewcommand\arraystretch{0.5}
  \centering
  \begin{adjustbox}{max width=.93\textwidth}
  \setlength{\tabcolsep}{0.9mm}{
  \begin{tabular}{|Cl|Cc|Cc|Cc|Cc|Cc|Cc|Cc|Cc|Cc|Cc|Cl|Cc|Cc|Cc|Cc|Cc|Cc|Cc|Cc|}
    \hline
    \footnotesize \textbf{7-Scenes} & \multicolumn{2}{c|}{\footnotesize DSAC++~\cite{Brachmann_2018_CVPR}}& \multicolumn{2}{c|}{\footnotesize AS~\cite{SattlerLK17}} &\multicolumn{2}{c|}{\footnotesize Inloc~\cite{inloc}} & \multicolumn{2}{c|}{\footnotesize Regression-only} &\multicolumn{2}{c|}{\footnotesize Ours} & \footnotesize \textbf{12-Scenes} & \multicolumn{2}{c|}{\footnotesize DSAC++~\cite{Brachmann_2018_CVPR}}  & \multicolumn{2}{c|}{\footnotesize Regression-only} &\multicolumn{2}{c|}{\footnotesize Ours} \\
    \hline
    \footnotesize ||&\footnotesize \textbf{Acc.} & \footnotesize \textbf{Med. Err.}   &  \footnotesize \textbf{Acc.} & \footnotesize \textbf{Med. Err.} &  \footnotesize \textbf{Acc.} & \footnotesize \textbf{Med. Err.} &  \footnotesize \textbf{Acc.} & \footnotesize \textbf{Med. Err.} &  \footnotesize \textbf{Acc.} & \footnotesize \textbf{Med. Err.} & \footnotesize ||& \footnotesize \textbf{Acc.} & \footnotesize \textbf{Med. Err.} &  \footnotesize \textbf{Acc.} & \footnotesize \textbf{Med. Err.} &  \footnotesize \textbf{Acc.} & \footnotesize \textbf{Med. Err.}\\
    \hline
    \footnotesize Chess      &  \footnotesize 97.1  & \footnotesize \textbf{0.02}, \textbf{0.5}  & \footnotesize - & \footnotesize 0.04, 2.0 & \footnotesize - & \footnotesize 0.03, 1.1 &\footnotesize 95.4  & \footnotesize \textbf{0.02}, 0.7 & \footnotesize \textbf{97.5} & \footnotesize \textbf{0.02}, 0.7    &\footnotesize Kitchen-1 & \footnotesize \textbf{100} & \footnotesize - &  \footnotesize \textbf{100} &\footnotesize \textbf{0.008}, \textbf{0.4} &\footnotesize \textbf{100}  & \footnotesize \textbf{0.008}, \textbf{0.4}\\
    
   \footnotesize Fire      & \footnotesize 89.6 & \footnotesize \textbf{0.02}, \textbf{0.9} & \footnotesize - &  \footnotesize  0.03, 1.5 & \footnotesize - &  \footnotesize  0.03, 1.1 & \footnotesize 94.9   & \footnotesize \textbf{0.02}, \textbf{0.9} & \footnotesize \textbf{96.7}  &\footnotesize \textbf{0.02}, \textbf{0.9}  &\footnotesize Living-1  & \footnotesize \textbf{100} & \footnotesize - & \footnotesize \textbf{100} &  \footnotesize \textbf{0.011}, \textbf{0.4}&\footnotesize \textbf{100}&\footnotesize \textbf{0.011}, \textbf{0.4}\\
   
   \footnotesize Heads       & \footnotesize 92.4 & \footnotesize \textbf{0.01}, \textbf{0.8} & \footnotesize - &  \footnotesize 0.02, 1.5 & \footnotesize - &  \footnotesize 0.02, 1.2 & \footnotesize 97.1   & \footnotesize \textbf{0.01}, \textbf{0.8} & \footnotesize \textbf{100}  &\footnotesize \textbf{0.01}, 0.9  &\footnotesize Bed  & \footnotesize 99.5 & \footnotesize - & \footnotesize \textbf{100} &  \footnotesize 0.013, 0.6&\footnotesize \textbf{100}&\footnotesize \textbf{0.009}, \textbf{0.4}\\
   
   \footnotesize Office       & \footnotesize \textbf{86.6} & \footnotesize \textbf{0.03}, \textbf{0.7} & \footnotesize - &  \footnotesize 0.09, 3.6 & \footnotesize - &  \footnotesize \textbf{0.03}, 1.1  & \footnotesize 81.4   & \footnotesize \textbf{0.03}, 0.9 & \footnotesize 86.5  &\footnotesize \textbf{0.03}, 0.8  &\footnotesize Kitchen-2  & \footnotesize 99.5 & \footnotesize - & \footnotesize \textbf{100} &  \footnotesize 0.008, 0.4&\footnotesize \textbf{100}&\footnotesize \textbf{0.007}, \textbf{0.3}\\
   
   \footnotesize Pumpkin       & \footnotesize 59.0  & \footnotesize  \textbf{0.04}, 1.1 & \footnotesize  - &  \footnotesize 0.08, 3.1 & \footnotesize  - &  \footnotesize 0.05, 1.6 & \footnotesize 58.0   & \footnotesize \textbf{0.04}, 1.1 & \footnotesize \textbf{59.9}  &\footnotesize \textbf{0.04}, \textbf{1.0}  &\footnotesize Living-2  & \footnotesize \textbf{100} & \footnotesize - & \footnotesize \textbf{100} &  \footnotesize 0.014, 0.6 &\footnotesize \textbf{100}&\footnotesize \textbf{0.010}, \textbf{0.4}\\
   
   \footnotesize Kitchen       & \footnotesize \textbf{66.6} & \footnotesize \textbf{0.04}, \textbf{1.1} & \footnotesize - &  \footnotesize 0.07, 3.4 & \footnotesize - &  \footnotesize \textbf{0.04}, 1.3  & \footnotesize 56.5   & \footnotesize 0.05, 1.4 & \footnotesize 65.5  &\footnotesize \textbf{0.04}, 1.2  &\footnotesize Luke  & \footnotesize 95.5 & \footnotesize - & \footnotesize 93.8 &  \footnotesize 0.020, 0.9&\footnotesize \textbf{96.3}&\footnotesize \textbf{0.012}, \textbf{0.5}\\
   
   \footnotesize Stairs       & \footnotesize 29.3 & \footnotesize 0.09, 2.6 & \footnotesize -&  \footnotesize \textbf{0.03}, 2.2 & \footnotesize -&  \footnotesize 0.09, 2.5  & \footnotesize 68.1   & \footnotesize 0.04, 1.0 & \footnotesize \textbf{87.5}  &\footnotesize \textbf{0.03}, \textbf{0.8}  &\footnotesize Gates 362  & \footnotesize \textbf{100} & \footnotesize - &  \footnotesize \textbf{100}&  \footnotesize 0.011, 0.5&\footnotesize \textbf{100}&\footnotesize \textbf{0.010}, \textbf{0.4}\\
   
    \cline{1-11}
    \footnotesize Average      & \footnotesize 74.4  &\footnotesize 0.04, 1.1  & \footnotesize - & \footnotesize 0.05, 2.5 & \footnotesize - & \footnotesize 0.04, 1.4  & \footnotesize  78.8   & \footnotesize  \textbf{0.03}, 1.0 & \footnotesize \textbf{84.8}   & \footnotesize \textbf{0.03}, \textbf{0.9} & \footnotesize Gates 381&  \footnotesize 96.8  & \footnotesize - & \footnotesize 98.8 & \footnotesize 0.016, 0.7& \footnotesize \textbf{99.1 }& \footnotesize \textbf{0.012}, \textbf{0.6}\\
    
   \cline{1-11}
    \footnotesize Complete     & \multicolumn{2}{c|}{\footnotesize 76.1}  & \multicolumn{2}{c|}{\footnotesize - } & \multicolumn{2}{c|}{\footnotesize - } & \multicolumn{2}{c|}{\footnotesize 74.7}  &\multicolumn{2}{c|} {\footnotesize \textbf{80.5}}  & \footnotesize Lounge&  \footnotesize 95.1 &\footnotesize - &\footnotesize 99.4 & \footnotesize 0.015, \textbf{0.5} &  \footnotesize \textbf{100}& \footnotesize \textbf{0.014}, \textbf{0.5}\\
   \cline{1-11}
   \footnotesize \textbf{Cambridge}  &\multicolumn{2}{c|}{\footnotesize  DSAC++~\cite{Brachmann_2018_CVPR}}&\multicolumn{2}{c|}{\footnotesize  AS~\cite{SattlerLK17}} &\multicolumn{2}{c|}{\footnotesize NG-RANSAC~\cite{Brachmann_2019_ICCV_NG}}&\multicolumn{2}{c|}{\footnotesize Regression-only} &\multicolumn{2}{c|}{\footnotesize Ours} & \footnotesize Manolis&  \footnotesize 96.4 & \footnotesize -  & \footnotesize 97.2   &\footnotesize 0.014, 0.7   &\footnotesize \textbf{100} &\footnotesize \textbf{0.011}, \textbf{0.5}\\
   \cline{1-11}
    \footnotesize Great Court       &\multicolumn{2} {c|}{\footnotesize 0.40, \textbf{0.2}}    &\multicolumn{2} {c|}{\footnotesize - }  &\multicolumn{2}{c|}{\footnotesize 0.35, -} &\multicolumn{2} {c|}{\footnotesize 1.25, 0.6} &\multicolumn{2}{c|} {\footnotesize \textbf{0.28}, \textbf{0.2}} &\footnotesize Floor5a & \footnotesize 83.7 &\footnotesize -  &\footnotesize 97.0 &\footnotesize 0.016, 0.7 &\footnotesize \textbf{98.8} &\footnotesize \textbf{0.012}, \textbf{0.5}\\
    
    \footnotesize K. College  &\multicolumn{2} {c|}{\footnotesize 0.18, \textbf{0.3}}    &\multicolumn{2} {c|}{\footnotesize 0.42, 0.6} &\multicolumn{2}{c|}{\footnotesize \textbf{0.13}, -} &\multicolumn{2} {c|}{\footnotesize 0.21, \textbf{0.3}} &\multicolumn{2}{c|} {\footnotesize 0.18, \textbf{0.3}}   &\footnotesize Floor 5b&  \footnotesize 95.0 & \footnotesize  -  &\footnotesize 93.3 &\footnotesize 0.019, 0.6 &\footnotesize \textbf{97.3}   & \footnotesize \textbf{0.015}, \textbf{0.5}\\
    \cline{12-18}
    \footnotesize Old Hospital    &\multicolumn{2} {c|}{\footnotesize 0.20, \textbf{0.3}}    &\multicolumn{2} {c|}{\footnotesize 0.44, 1.0  } &\multicolumn{2}{c|}{\footnotesize 0.22, -} &\multicolumn{2} {c|}{\footnotesize 0.21, \textbf{0.3}} &\multicolumn{2}{c|}{\footnotesize \textbf{0.19}, \textbf{0.3}}  &   \footnotesize Average&  \footnotesize 96.8 & \footnotesize -   &\footnotesize 98.3 &\footnotesize 0.014, 0.6 &\footnotesize \textbf{99.3} &\footnotesize \textbf{0.011}, \textbf{0.5}\\
    \cline{12-18}
    \footnotesize Shop Facade      &\multicolumn{2} {c|}{\footnotesize \textbf{0.06}, \textbf{0.3}}    &\multicolumn{2} {c|}{\footnotesize 0.12, 0.4  } &\multicolumn{2}{c|}{\footnotesize \textbf{0.06}, -} &\multicolumn{2} {c|}{\footnotesize \textbf{0.06}, \textbf{0.3}} & \multicolumn{2}{c|}{ \footnotesize \textbf{0.06}, \textbf{0.3}}  & \footnotesize Complete &\multicolumn{2} {c|}{\footnotesize 96.4}&\multicolumn{2} {c|}{\footnotesize 97.9}&\multicolumn{2} {c|}{\footnotesize \textbf{99.1}} \\
    \cline{12-18}
   \footnotesize St M. Church      &\multicolumn{2} {c|}{\footnotesize 0.13, 0.4}    &\multicolumn{2} {c|}{\footnotesize 0.19, 0.5 }  &\multicolumn{2}{c|}{\footnotesize 0.10, -} &\multicolumn{2} {c|}{\footnotesize 0.16, 0.5} &\multicolumn{2}{c|}{\footnotesize \textbf{0.09}, \textbf{0.3}} \\
   \cline{1-11}
    \footnotesize Average     &\multicolumn{2} {c|}{\footnotesize 0.19, \textbf{0.3}}    &\multicolumn{2} {c|}{\footnotesize 0.29, 0.6 }  &\multicolumn{2}{c|}{\footnotesize 0.17, -} &\multicolumn{2} {c|}{\footnotesize 0.38, 0.4} &\multicolumn{2}{c|}{\footnotesize \textbf{0.16}, \textbf{0.3}} \\
   \cline{1-11}
  \end{tabular}}
   \end{adjustbox}
  \vspace{1ex}
  \caption{The median errors (m, $^\circ$) for 7-Scenes, 12-Scenes and Cambridge, and the percentages of accurately localized test images (error $< 5 \textrm{ cm}, 5^\circ$) for 7-Scenes and 12-Scenes. 
  ``Complete'' refers to the percentage among all test images of all scenes. 
  \label{tab:eval:loc}}%
 \vspace{-2mm}
\end{table*}

\subsection{Datasets and Experimental Setup}\label{sec:setup}

We use four standard benchmark datasets for our experiments. 
The \textbf{7-Scenes} (7S)~\cite{SCoRF} dataset is a widely used RGB-D dataset that contains seven indoor scenes. RGB-D image sequences of the scenes are recorded by a KinectV1. Ground truth poses and dense 3D models are also provided.
 \textbf{12-Scenes} (12S)~\cite{valentin2016learning}  is another  indoor RGB-D dataset. It is composed of twelve rooms captured with a Structure.io depth sensor and an iPad color camera, and ground truth poses are provided along with the RGB-D images. The recorded environments are significantly larger than those in 7-Scenes. 
  \textbf{Cambridge Landmarks}~\cite{kendall2015convolutional}   is an outdoor RGB visual localization dataset. It consists of RGB images of six scenes captured using a Google LG Nexus 5 smartphone. Ground truth poses and sparse 3D reconstructions generated with structure from motion are also provided.
In addition to these three datasets, we synthesize three large-scale indoor scenes based on 7-Scenes and 12-Scenes by placing all seven, twelve or nineteen individual scenes, into a single coordinate system similar to~\cite{Brachmann_2019_ICCV}. These large integrated datasets are denoted by \textbf{i7-Scenes} (i7S), \textbf{i12-Scenes} (i12S), \textbf{i19-Scenes} (i19S), respectively.  Finally, we evaluate our method on the \textbf{Aachen Day-Night} dataset~\cite{sattler2018benchmarking,sattler2012image} %
which is very challenging for scene coordinate regression methods due to the scale and sparsity of the 3D model. In addition, it contains a set of challenging night time queries, but there is no night time training data. In the following, we present the main setup for experiments on all the datasets except Aachen. See supplementary for details on Aachen.  

Ground truth scene coordinates can be either obtained from the known poses and depth maps or  rendered using a 3D model. To generate the ground truth location labels,  we run hierarchical k-means clustering on dense point cloud models. 
For all the individual scenes used in the main experiments, unless stated otherwise,  we use two-level hierarchical k-means with the branching factor set to 25 for both levels. 
For the three combined scenes, {i7-Scenes}, {i12-Scenes}, and {i19-Scenes}, we simply combine the label trees at the first level. That is, \eg, for the i7-Scenes, there are 175 branches in total at the first level.%

We use the same VGG-style~\cite{simonyan2014very} architecture as DSAC++~\cite{Brachmann_2018_CVPR} as the base regression network for our method, except we use ELU activation~\cite{ClevertUH15} instead of ReLU~\cite{nair2010rectified}. This is because we found that the plain regression network is easier to train with ReLU, while our network which has the additional conditioning layers and classification branches works better with ELU.  
The regression layer, the second and first classification layer have a receptive field size of 73$\times$73, 185$\times$185, and 409$\times$409 pixels, respectively, in the input image.
 To show the advantage of the proposed architecture, we also evaluate the localization performance of the same regression-only network used in DSAC++~\cite{Brachmann_2018_CVPR}, but here trained with the Euclidean loss term only. Note that in~\cite{Brachmann_2018_CVPR},  two additional training  steps are  proposed and the entire localization pipeline is optimized end-to-end, which can further improve the accuracy. Potentially, our network can  also benefit from the DSAC++ framework, but it is beyond the scope of the current paper.  
Unless specified otherwise, we perform affine data augmentation with additive brightness changes during training. We also report the results obtained without data augmentation in \sect{ablation}. %
For pose estimation, we follow~\cite{Brachmann_2018_CVPR}, and  use the same PnP-RANSAC algorithm with the same hyperparameter settings.
 Further details about the architecture, training and other settings can be found in the supplementary material.

\subsection{Results on 7-Scenes, 12-Scenes and Cambridge}

To evaluate our hierarchical joint learning architecture, we first compare it with the state-of-the-art methods as well as a regression-only baseline on the 7-Scenes, the 12-Scenes, and the Cambridge Landmarks datasets. For the Cambridge Landmarks, we report median pose accuracy as in the previous works. Following~\cite{Brachmann_2018_CVPR,Brachmann_2019_ICCV_NG,Li_Ylioinas_Verbeek_Kannala_2018}, we do not include the Street scene, since the dense 3D reconstruction of this scene has rather poor  quality that hampers performance. For the 7-Scenes and the 12-Scenes, we also report the percentage of the test images with error below 5 cm and 5$^\circ$, which is used as the main evaluation metric for both datasets and gives more information about the localization performance. 
Scene coordinate regression methods are currently the best performing single-image RGB methods on these three small/medium scale datasets~\cite{Brachmann_2018_CVPR, Brachmann_2019_ICCV_NG}. 
We also compare to a state-of-the-art feature-based method, \ie Active Search~\cite{SattlerLK17} and an indoor localization method which exploits dense correspondences~\cite{inloc}. Note that, in general, methods that exploit additional depth information~\cite{Cavallari_corr_19,cavallari2019real} or sequences of images~\cite{radwan2018vlocnet++,valada2018deep} can provide better localization performance. However, the additional required information  also restricts the scenarios in which they can be applied. 
We do not compare to those methods in this work, since they are not directly comparable to our results produced in the single-image RGB localization setting.

The results are reported in \tab{eval:loc}. Numbers for the competing methods are taken from the corresponding papers.
Overall, our approach yields excellent results. Compared to the  regression-only baseline, our approach provides consistently better localization performance on all the scenes across the three datasets.
During training, we also observed consistently lower regression training error compared to the regression-only baseline, underlining the ability of the discrete location labels to disambiguate the local appearances.
Our approach also achieves overall better results compared to the current state-of-the-art methods DSAC++~\cite{Brachmann_2018_CVPR} on all three datasets, and NG-RANSAC~\cite{Brachmann_2019_ICCV_NG} on the Cambridge Landmarks (the latter does not report results on the 7-Scenes and 12-Scenes datasets).
In \tab{eval:loc} we trained our networks and the  regression-only baseline with data augmentation, while DSAC++ and NG-RANSAC did not use  data augmentation. 
In \sect{ablation}, we show that even without data augmentation, our method still achieves comparable or better performance compared to DSAC++ and NG-RANSAC. Moreover, in DSAC++ and NG-RANSAC, more advanced training  steps and RANSAC schemes are proposed to improve the accuracy of the plain regression network and to optimize the entire pipeline, while in this work we  focus on the scene coordinate network itself and we show that improvements on  this single component can already  improve  the localization performance beyond the state-of-the-art. 
Note that DSAC++ and NG-RANSAC are complementary to our approach, and their combination could be explored in future work.

\begin{figure}
     \begin{center}
     \includegraphics[width=1\linewidth]{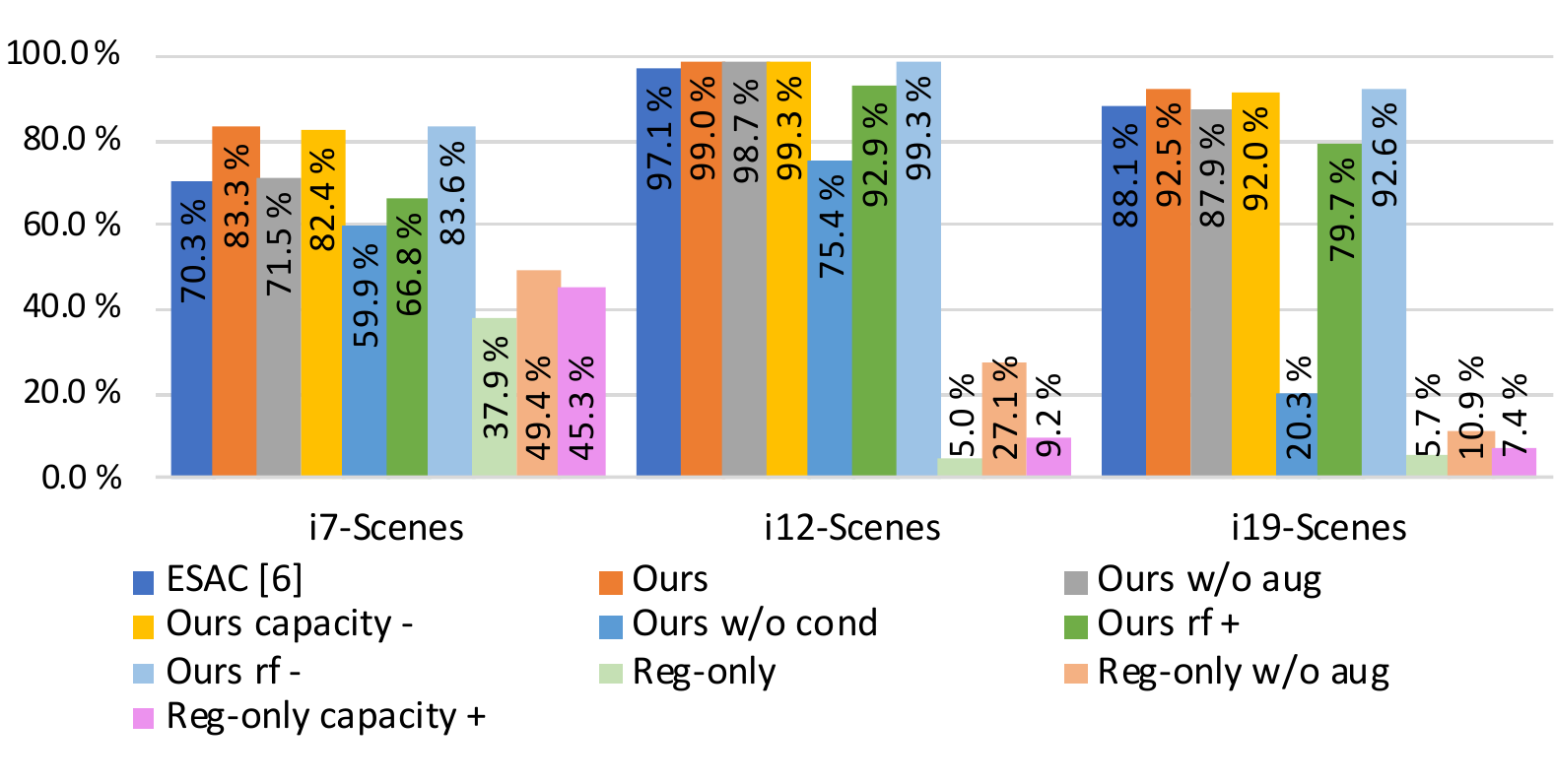}
     \end{center}
     \caption{Average pose accuracy on the combined scenes. 
     Results for ESAC taken from~\cite{Brachmann_2019_ICCV}. Our method consistently outperforms the regression-only baseline by a large margin and achieves better performance compared to ESAC.
     }
     \label{fig:i12i19}
\end{figure}

\subsection{Results on Combined Scenes}
The individual scenes from the previous datasets all have very limited physical extent. As in~\cite{Brachmann_2019_ICCV}, to go beyond such small environments, we use the combined scenes, \ie the i7-Scenes, i12-Scenes, and the i19-Scenes, as described in \sect{setup}. We mainly compare to the regression-only baseline and ESAC~\cite{Brachmann_2019_ICCV} on the three combined scenes. To the best of our knowledge, ESAC is currently the only scene coordinate regression method that scales well to the combined scenes. The results are reported in \fig{i12i19}. 

We see that the localization performance of the regression baseline (\textit{Reg-only}) decreases dramatically when trained on the combined scenes compared to trained and tested on each of the scenes individually, \cf \tab{eval:loc}. Its performance drops more drastically as the scene grows larger.
Our method is much more robust to the increase in the environment size, and  significantly outperforms  the baseline.
This underlines the importance of our hierarchical learning framework
when the environment is large and potentially contains more ambiguities. 
Our method also outperforms ESAC which uses an ensemble of networks, where
each network specializes in a local part of the environment~\cite{Brachmann_2019_ICCV}. ESAC requires to train and store  multiple networks, whereas our approach requires only a single model. 

Note that for ESAC the authors did not use data augmentation. When we train our method without data augmentation (\textit{Ours w/o aug}), we  still outperform ESAC on i7-Scenes and i12-Scenes, and obtain a slightly lower but comparable accuracy on i19-Scenes (87.9\% \vs 88.1\%).
Note that ESAC and our approach are complementary, and their combination could be explored in future work.

\begin{table}
\centering
\renewcommand\arraystretch{0.1}
 
 \begin{adjustbox}{max width=0.95\linewidth}
 \setlength{\tabcolsep}{1.5mm}{
 \begin{tabular}{cccccc}
 \toprule
  \footnotesize Reg-only & \footnotesize Ours & \footnotesize Ours capacity-  &\footnotesize ESAC (i7S)~\cite{Brachmann_2019_ICCV} &\footnotesize ESAC (i12S)~\cite{Brachmann_2019_ICCV} &\footnotesize ESAC (i19S)~\cite{Brachmann_2019_ICCV} \\
 \midrule
   \footnotesize 104MB & \footnotesize 165MB & \footnotesize 73MB &\footnotesize 7$\times$28MB &\footnotesize 12$\times$28MB &\footnotesize 19$\times$28MB \\

 \bottomrule
 \end{tabular}}
  \end{adjustbox}
 \vspace{1ex}
 \caption{Model size comparison. Our method can scale robustly to large environments with a compact model.
 \label{tab:modelsize}}
 \vspace{-2mm}
\end{table}

\subsection{Detailed Analysis} \label{sec:ablation}
\mypar{Network capacity} 
Compared to the regression-only baseline, our network has extra layers for the conditioning generators and classification branches, and thus has an increased number of parameters.  Therefore, for fair comparison, we add more channels to the  regression-only baseline to compensate the increased number of parameters in our model. On 7-Scenes, the average accuracy of the regression baseline increased from $78.8\%$ to $80.4\%$. On the combined scenes, as shown in \fig{i12i19}, we observe larger improvement in performance (denoted by \textit{Reg-only~capacity+} in \fig{i12i19}). However, even with increased capacity, the  regression-only baseline still lags far behind our method, especially on the combined scenes. 

We also experimented with reducing the size of the backbone regression network, which accounts for most of the model parameters.  
We add more conditioning layers early in the network, while  using less shared layers between the regression and classification branches. 
We denote the resulting network by \textit{Ours~capacity-}, see supplementary for details.
In \tab{modelsize}, we compare the model size of our network to the regression baseline and ESAC on the combined scenes. 
We  see in \fig{i12i19} and \tab{modelsize} that this allows us to reduce our  model size  by more than a factor of two, while incurring a loss in  accuracy below one percentage point. 
Compared to ESAC on the i19-Scenes dataset, our compressed model is more than seven times more compact.
Note that since we perform regression locally, the k-means cluster centers also need to be stored. Since for each individual scene there are only 625 clusters, the storage space needed for the cluster centers is negligible ($<$ 1MB).

\mypar{Using global information}
Using global information directly to regress scene coordinates has been explored in~\cite{Li2018}. However, even with data augmentation, large input patterns remain sensitive to viewpoint changes, leading to inferior performance at test time compared to using local patches~\cite{Brachmann_2018_CVPR}. We validate this by using the same  regression network, but now with dilated convolution such that the receptive field size is much larger (409$\times$409). We find that in general directly using global context helps the training loss decrease faster. 
This might have a positive effect on complex scenes ($39.3\%$ with dilated convolution \vs $37.9\%$ without it on i7-Scenes). For less demanding scenes, however, the network usually gives worse results ($59.2\%$ \vs $78.8\%$ on 7-Scenes) due to decreased viewpoint invariance.
Meanwhile, our network is able to use the global information in a more robust way, \ie, indirectly through discrete location labels. 

We also created two variants of our network  with small (73$\times$73)  and  large (409$\times$409) receptive field  across all levels, denoted by \textit{Ours~rf-} and \textit{Ours~rf+} respectively in \fig{i12i19}. As expected, increasing the receptive field size at all levels harms the performance,  as shown in \fig{i12i19}. Interestingly, the model with small receptive field even performs sightly better on the combined scenes. This indicates that the local ambiguities can be handled well by the hierarchical coarse-to-fine conditioning mechanism. 

\begin{table}
\centering
\renewcommand\arraystretch{0.1}
 \begin{adjustbox}{max width=0.9\linewidth}
 \setlength{\tabcolsep}{4mm}{
 \begin{tabular}{ClCcCcCc}
 \toprule
   &\footnotesize 7-Scenes & \footnotesize 12-Scenes & \footnotesize Cambridge \\
 \midrule
 \footnotesize Reg-only w/o aug & \footnotesize 70.9\% & \footnotesize 97.5\% & \footnotesize 0.38m, 0.4\textdegree \\
 \footnotesize Ours w/o aug& \footnotesize 75.5\% & \footnotesize 99.4\% & \footnotesize 0.18m, 0.3\textdegree \\
  \footnotesize DSAC++~\cite{Brachmann_2018_CVPR} & \footnotesize 74.4\%  & \footnotesize 96.8\%   & \footnotesize 0.19m, 0.3\textdegree{}   \\
 \footnotesize  NG-RANSAC~\cite{Brachmann_2019_ICCV_NG}& \footnotesize - & \footnotesize  -  & \footnotesize  0.17m, -  \\
 \bottomrule
 
 \end{tabular}}
  \end{adjustbox}
 \vspace{1ex}
 \caption{Average pose accuracy/median error on the 7-Scenes, 12-Scenes and Cambridge datasets of our method and the regression-only baseline without data augmentation.}
  \label{tab:eval:aug}
  \vspace{-2mm}
\end{table}

\mypar{Data augmentation} 
We apply  affine transformations to the images with additive brightness changes as data augmentation during training. In general, this improves the generalization capability of the network and makes it more robust to lighting and viewpoint changes. According to \tab{eval:loc}, \tab{eval:aug} and \fig{i12i19}, data augmentation consistently improves the localization performance of our method, except on the 12-Scenes dataset; in 12S, the training and test trajectories are close, and there are no significant viewpoint changes between training and test frames~\cite{cavallari2019real}. 
Data augmentation, however, can also increase the appearance ambiguity of the training data and make the network training more difficult. This happens to the baseline regression-only network: Although data augmentation helps it on the small-scale scenes, on the Cambridge and the combined scenes, data augmentation has no positive effects and even harms the performance.  
Note that without data augmentation, our method  still provides results that are better than or on par  with  the state-of-the-arts, see \tab{eval:aug} and \fig{i12i19}. 

\mypar{Conditioning  mechanism}
By formulating the scene regression task  as a coarse-to-fine joint classification-regression task can help
break the complexity of the original regression problem to some extent, even without the proposed conditioning mechanism. 
To show this experimentally, we trained a variant of our network without the conditioning  mechanism, \ie we removed all the conditioning generators and layers, thus no coarse location information is fed to influence the network activations at the finer levels.
We did  preserve the coarse-to-fine joint learning, and still use the predicted location labels to determine the k-means cluster \wrt which the local regression coordinates are predicted.
We denote this model variant by \textit{Ours~w/o~cond}. 
 In contrast to the regression-only baseline, the regression part still learns to perform local regression by predicting the offsets with respect to the cluster centers of the finest classification hierarchy.
As shown in \fig{i12i19}, this variant outperforms the regression-only baseline, and significant performance gain can be observed on the combined scenes. However, compared to our full architecture, it still falls far behind, especially on the largest i19-Scenes. This illustrates that the proposed conditioning mechanism plays a crucial role in our hierarchical coarse-to-fine scene coordinate learning framework, and the significantly improved performance compared to the regression-only baseline is not achievable without it.

\begin{table}
\renewcommand\arraystretch{0.2}
  \centering
  \begin{adjustbox}{max width=0.9\linewidth}
  \setlength{\tabcolsep}{1mm}{
  \begin{tabular}{l|CcCcCcCcCcCcCc}
    \hline
    \multirow{2}{*}[3pt]{\footnotesize \textbf{7S}} &
    \footnotesize 9$\times$9 & \footnotesize 49$\times$49&\footnotesize 10$\times$100$\times$100 &\footnotesize 10$\times$100$\times$100$\times$100 &\footnotesize 625 &\footnotesize 25$\times$25\\
    \cline{2-7}
        & \footnotesize 82.9\% & \footnotesize 85.0\% &\footnotesize 85.9\% &\footnotesize 85.5\% &\footnotesize 85.3\% &\footnotesize 84.8\%\\
    \hline
    \hline
       \multirow{ 2}{*}[3pt]{\footnotesize \textbf{i7S}}& 
       \footnotesize 63$\times$9 &
       \footnotesize 343$\times$49&\footnotesize 70$\times$100$\times$100 &\footnotesize 70$\times$100$\times$100$\times$100 &\footnotesize 7$\times$25$\times$25 &\footnotesize 175$\times$25\\
    \cline{2-7}
       & \footnotesize 80.6\% & \footnotesize 83.7\% &\footnotesize 83.0\% &\footnotesize 82.1\% &\footnotesize 83.0\% &\footnotesize 83.3\%\\
    \hline
    
  \end{tabular}}
  \end{adjustbox}
  \vspace{1ex}
  \caption{Average pose accuracy obtained with different hierarchy settings. 
  The models with  4-level label hierarchy are classification-only, \ie the final  regression layer is omitted.
  \label{tab:eval:ll}}%
  \vspace{-2mm}
\end{table}

\mypar{Hierarchy and partition granularity}
In \tab{eval:ll} we report results obtained on the 7-Scenes and i7-Scenes datasets using  label hierarchies of different depth and width.
The results show that the performance of our approach is  robust \wrt the choice of these hyperparameters,  and only for the smallest 2-level label hierarchies that we tested we observed a significant drop in performance. Note that for the default setting (25$\times$25),  the results on 7-Scenes reported in Table \ref{tab:eval:loc} and \ref{tab:eval:ll} are the best across 10 runs of the randomly initialized k-means ($\text{mean}=84.3\%$, $\text{SD}= 0.4\%$). How to optimally partition the scene could be explored in future work.
%

\subsection{Outdoor Aachen Localization Results}\label{sec:aachen}
\begin{figure}[t]
\begin{center}

   \includegraphics[width=0.95\linewidth]{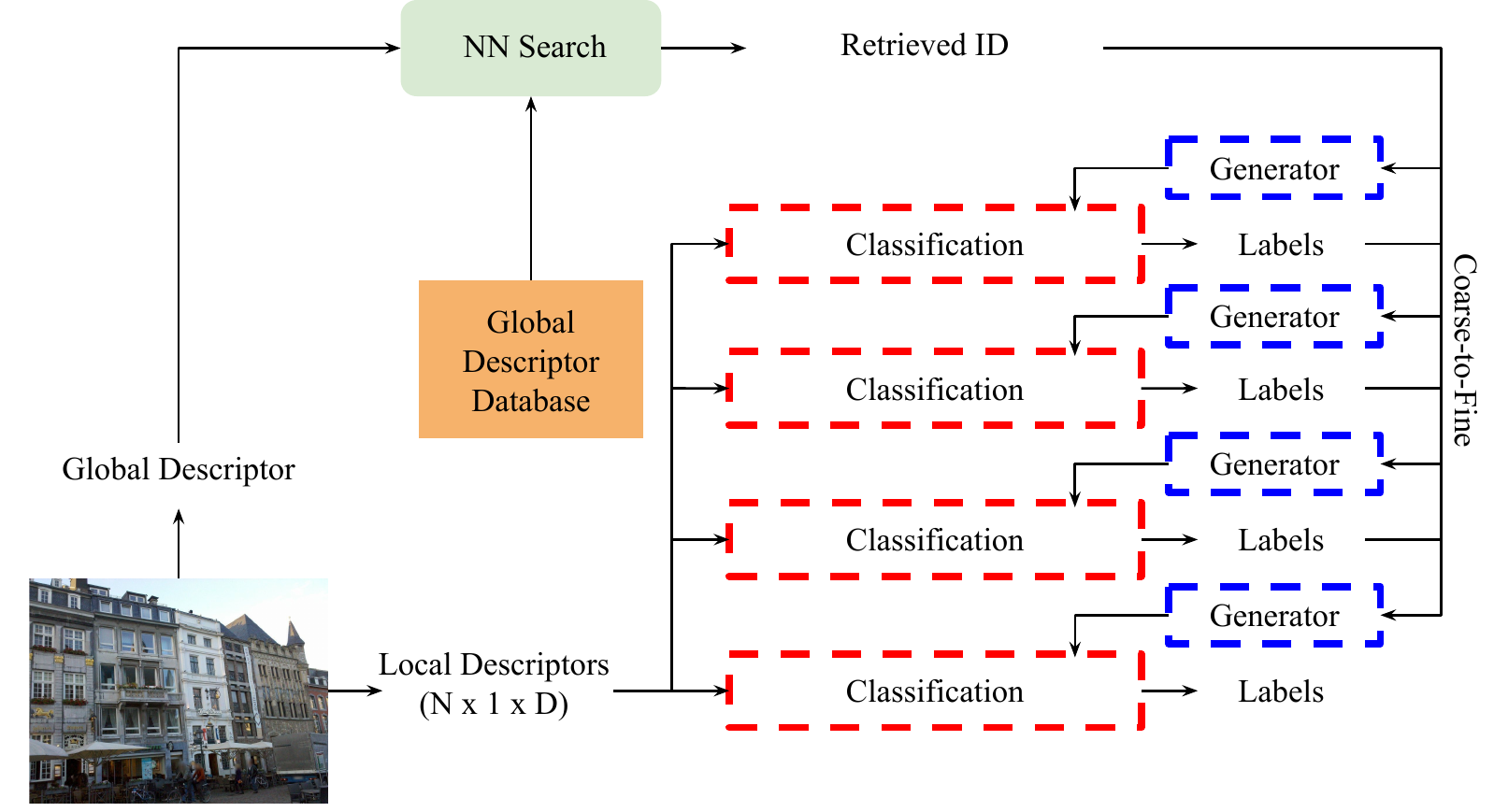}
\end{center}
   \caption{Illustration of our method with sparse local features and global image retrieval used in  the  Aachen dataset experiments. 
\label{fig:aachen}}
\end{figure}
The Aachen dataset is a challenging outdoor large-scale dataset, which is particularly  difficult for scene coordinate regression methods duo to the lack of dense model, the city-scale environment, and the night time queries. 
To the best of our knowledge, ESAC is the only existing method of this kind which  gives reasonable results on this dataset. 

We present a hybrid approach built upon our network for the challenging dataset.
To resolve the sparsity of the training data, in~\cite{Brachmann_2019_ICCV}, a re-projection error~\cite{Brachmann_2018_CVPR,Li_Ylioinas_Verbeek_Kannala_2018} is optimized densely, which is not applicable to our method. Therefore, we resort to sparse local features~\cite{DeTone_2018_CVPR_Workshops,Dusmanu2019CVPR}, such that during both training and test, our network only takes in a list of sparse features as input rather than a dense RGB image. 
To use image-level contextual information, we adopt an  image retrieval technique.
In addition to the location labels, every output layer including the first one is also conditioned on an image ID. During training, it is the ID of the training image. At test time, it is the ID of a retrieved image. 
We use SuperPoint~\cite{DeTone_2018_CVPR_Workshops} as the local feature, and NetVLAD~\cite{NetVLAD} for global image retrieval. 
The results in \tab{aachen} show that  for the Aachen dataset   the classification-only variant performs better, although it is not always the case, see \tab{eval:ll}. 
We use a 4-level classification-only network, and at the finest level, each cluster contains only one single 3D point. 
We use the retrieved database image   also  to perform a simple pre-RANSAC filtering step. Since the predictions are conditioned on the image ID, a prediction that is not visible in the corresponding image is likely to be a false match. Therefore, we filter out the predictions that are not visible in the corresponding retrieved image before the RANSAC stage. As shown in \tab{aachen}, this further improves the performance. Since the top-1 image can be a false positive, we run the pipeline for all the top-10 images, and select the prediction with the largest number of inliers. 
See the supplementary material for more details.

This approach significantly outperforms ESAC, and its performance is comparable to Active Search. However, compared to the hierarchical localization method of~\cite{sarlin2019coarse} which  also uses SuperPoint and NetVLAD, our method still falls behind. Nevertheless, our method requires no database of local descriptors and the model size of our hierarchical network is 179MB, while in~\cite{sarlin2019coarse}, a  local descriptor database of 4GB is used. 
Our results reduce the gap between scene coordinate learning approaches and  the state-of-the-art feature-based methods on this dataset, and we expect our method to perform better if a dense model is available. An advantage of the scene coordinate learning methods is that the model size does not grow linearly with the number of points in the scene model. This allows these methods to implicitly and efficiently store a dense descriptor point cloud in the network, and to produce dense matches  at test time, which often leads to better pose estimation than sparse matches~\cite{inloc}.

\begin{table}[!t]
\centering
\renewcommand\arraystretch{0.2}

 \begin{adjustbox}{max width=.9\linewidth}
 \setlength{\tabcolsep}{1mm}{
 \begin{tabular}{ClCc|Cc}
 \toprule
  \multirow{2}{*}[3pt]{\footnotesize \textbf{Method}}  & \footnotesize Aachen Day & \footnotesize Aachen Night \\
  &
\scriptsize 0.25m, 2\textdegree{} / 0.5m, 5\textdegree{} / 5m, 10\textdegree{}  & \scriptsize 0.5m, 2\textdegree{} / 1m, 5\textdegree{} / 5m, 10\textdegree{} \\
 \midrule
\footnotesize AS~\cite{SattlerLK17} & \footnotesize 57.3\% / 83.7\% / \textbf{96.6\%} & \footnotesize 19.4\% / 30.6\% / 43.9\% \\
\footnotesize HL SP+NV~\cite{sarlin2019coarse} & \footnotesize \textbf{80.5\% }/ \textbf{87.4\% }/ 94.2\% & \footnotesize \textbf{42.9\%} / \textbf{62.2\%} / \textbf{76.5\%} \\
\footnotesize ESAC (50 experts)~\cite{Brachmann_2019_ICCV} & \footnotesize 42.6\% / 59.6\% / 75.5\% & \footnotesize 3.1\% / 9.2\% / 11.2\% \\
\footnotesize Ours top-10 w/ filt & \footnotesize 71.1\% / 81.9\% / 91.7\% & \footnotesize 32.7\% / 43.9\% / 65.3\% \\
 \midrule
 \footnotesize Ours top-10 w/ filt w/o aug & \footnotesize 65.5\% / 77.3\% / 88.8\% & \footnotesize 22.4\% / 38.8\% / 54.1\% \\
 \footnotesize Ours top-1 w/ filt & \footnotesize 64.0\% / 76.1\% / 85.4\% & \footnotesize 18.4\% / 32.7\% / 53.1\% \\
 \footnotesize Ours top-1  & \footnotesize 58.3\% / 66.4\% / 80.2\% & \footnotesize 13.3\% / 21.4\% / 32.7\% \\
  \footnotesize Ours w/o retreived ID& \footnotesize 50.6\% / 56.3\% / 70.1\% & \footnotesize 7.1\% / 11.2\% / 19.4\% \\
 \footnotesize Ours top-1 (4-level cls-reg) & \footnotesize 47.8\% / 61.8\% / 79.9\% & \footnotesize 10.2\% / 21.4\% / 35.7\% \\
 \footnotesize Ours top-1 (3-level cls-reg) & \footnotesize 20.9\% / 42.2\% / 76.9\% & \footnotesize 3.1\% / 14.3\% / 32.7\% \\
 \bottomrule
 \end{tabular}}
 \end{adjustbox}
 \vspace{1ex}
 \caption{Accuracy on the Aachen  dataset. Unless stated otherwise,  we use a 4-level classification-only network for our method.
 \label{tab:aachen}}
 \vspace{-2mm}
\end{table}    

\section{Conclusion}
\label{sec:conclusion}

We have proposed a novel  hierarchical coarse-to-fine  scene coordinate learning approach, enabled by a FiLM-like conditioning mechanism, for visual localization. Our network has several levels of output layers with each of them conditioned on the outputs of the previous ones. Progressively finer  localization labels are predicted  with classification branches. The scene coordinate predictions can be obtained through a final regression layer or using the cluster centers at the finest level.  
The results
show that the hierarchical scene coordinate  network leads to more accurate  camera re-localization performance than the previous regression-only approaches, achieving state-of-the-art results for single-image RGB localization on three benchmark datasets. Moreover, our novel architecture allows us to train  compact models which scale robustly to large environments, achieving state-of-the-art on three combined scenes. Finally, we show a hybrid approach that further narrows the gap to the state-of-the-art feature-based methods for challenging large-scale outdoor localization. 

{\footnotesize
\mypar{Acknowledgements}
This work has been supported by the Academy of Finland
(grants 277685, 309902), and the French National Research Agency (grants ANR16-CE23-0006, ANR-11-LABX0025-01). We acknowledge the computational resources provided by the Aalto Science-IT project and CSC ${\text -}$ IT Center for Science, Finland.
}
\newpage
{\small
\bibliographystyle{ieee_fullname}
\bibliography{bib}
}
\newpage
\appendix

\section*{\fontsize{18}{15}\selectfont
---Supplementary Material---}


In this supplementary material, we provide more details on network architecture, training procedure,  and other experimental settings.  Additional qualitative results are presented at the end. 

\section{Main Experiment Details}

In this section, we present the experiment details on 7-Scenes, 12-Scenes, Cambridge Landmarks, and the combined scenes.
\subsection{Network Architecture}


We use a similar VGG-style~\cite{simonyan2014very} architecture as DSAC++~\cite{Brachmann_2018_CVPR} as the base regression network, except we use ELU activation~\cite{ClevertUH15} instead of ReLU~\cite{nair2010rectified}. As mentioned in the main paper, we found that the plain regression network is faster to train with ReLU, while our network which has additional conditioning layers and classification branches works better with ELU. 
Conditioning layers and generators, as well as two additional classification branches, are added upon the base network for our 3-level hierarchical network which is used in the main experiments. 

There are  three convolutional layers with stride 2 in the regression base network.  The output resolution  of the regression branch is thus reduced by a factor of 8. Strided convolution, dilated convolution and upconvolution are used in the two classification branches to enlarge the receptive field and preserve the output resolution.
The predicted classification labels are converted into one-hot format before being fed into the generators. If more than one label map used as input to a conditioning generator, the label maps are concatenated.

\begin{figure*}[ht]
  \includegraphics[width=0.85\linewidth]{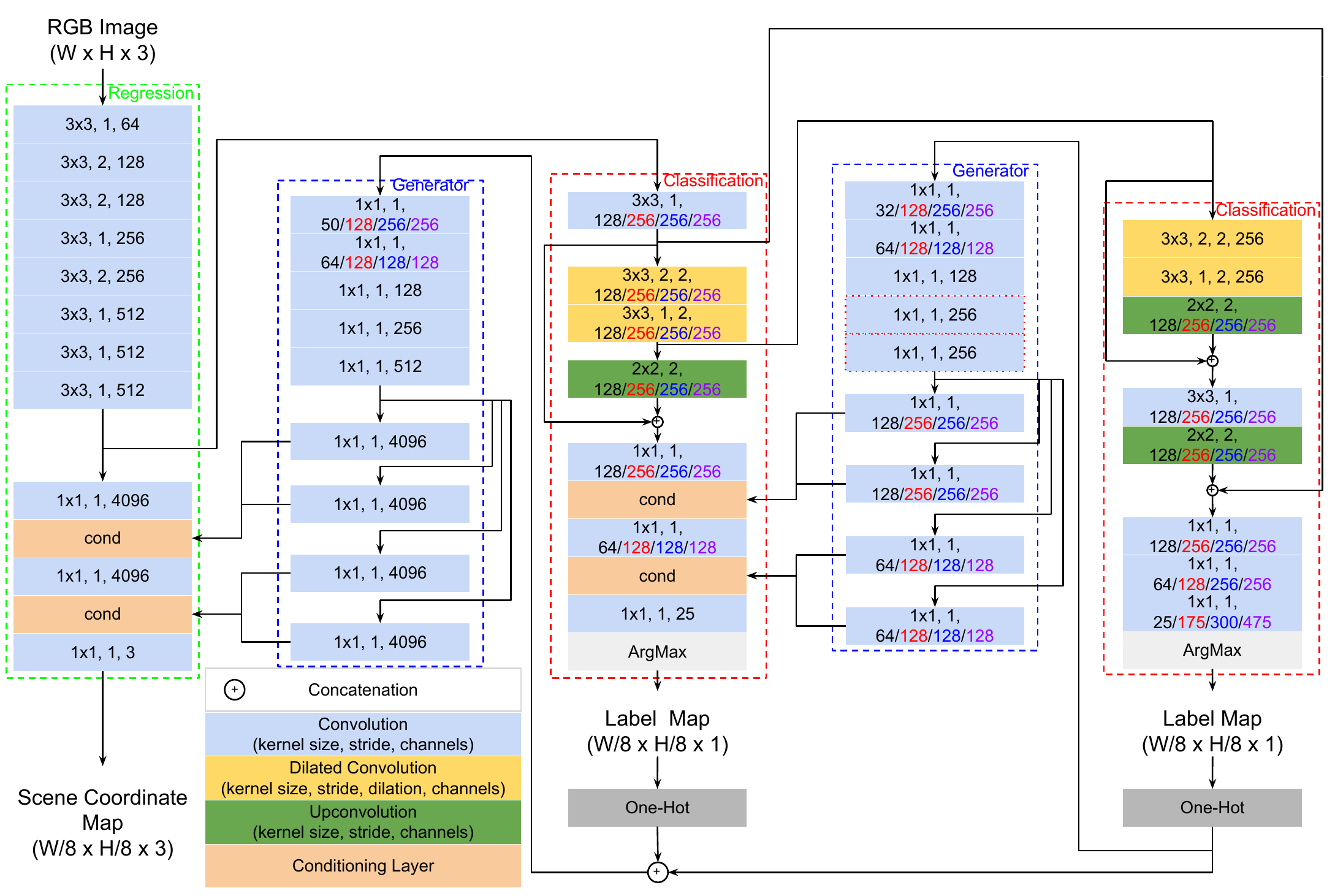}
   \centering
  \caption{Detailed network architecture.}
  \label{fig:net1}
\end{figure*}

\begin{figure*}[ht]
  \includegraphics[width=0.85\linewidth]{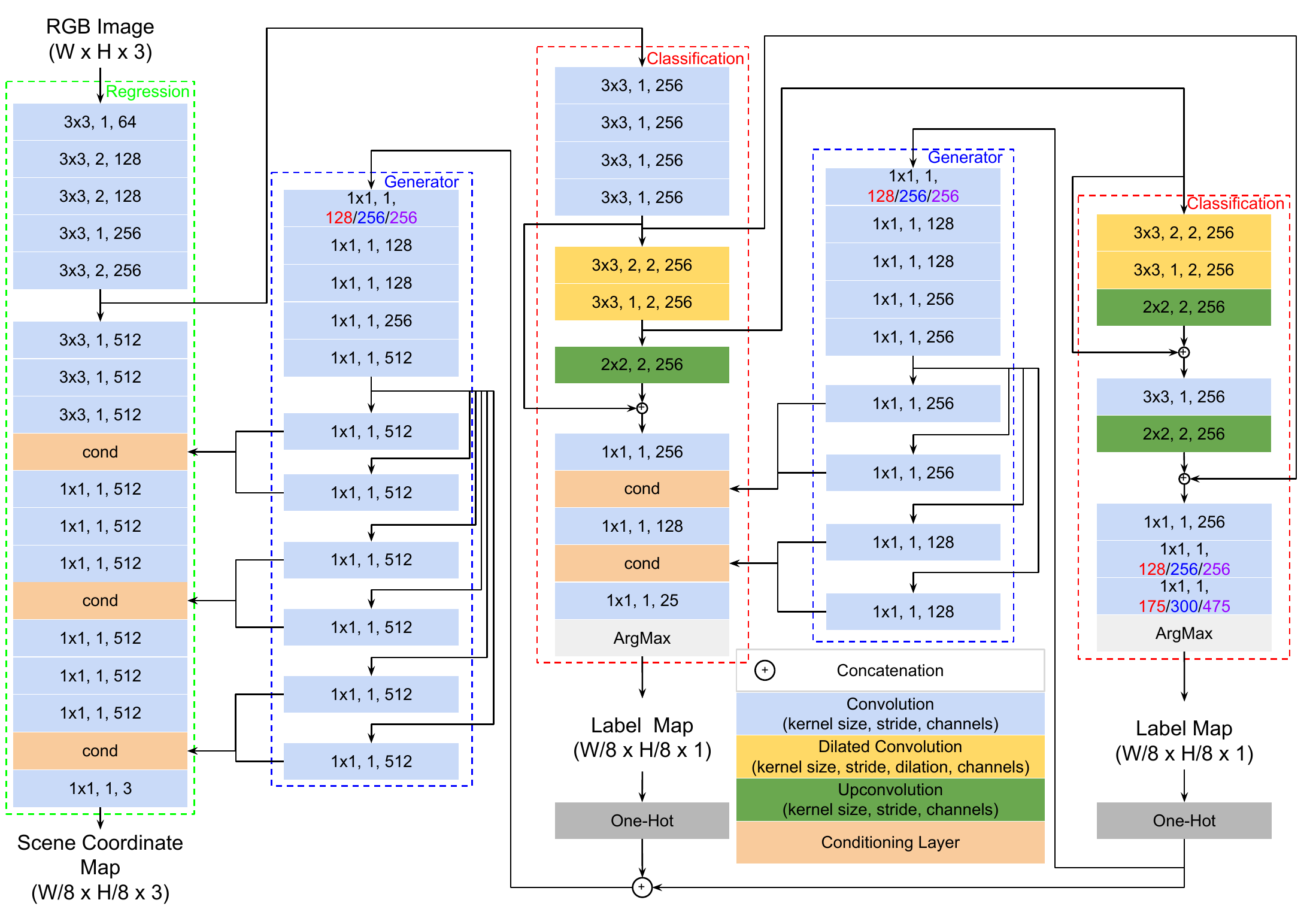}
   \centering
  \caption{The more compact architecture of \textit{Ours~capacity-}.}
  \label{fig:net2}

\end{figure*}

The detailed architecture is given in \fig{net1}. 
For experiments on 7-Scenes, 12-Scenes, Cambridge Landmarks, we use the same network architecture. 
For experiments on the combined scenes, we  increased the number of channels for certain layers and added two more layers in the first conditioning generator. 
The additional layers are marked in red, and the increased channel counts are marked in red, blue and purple for i7-Scenes, i12-Scenes and i19-Scenes, respectively. 
The more compact architecture for the experiments (\textit{Ours~capacity-} in \tab{modelsize} and \fig{i12i19} of the main paper) on the combined scenes is illustrated in \fig{net2}. In the case we use different numbers of channels for a convolutional layer,  the channel counts are marked in red, blue and purple for i7-Scenes, i12-Scenes and i19-Scenes respectively. 

As in DSAC++~\cite{Brachmann_2018_CVPR}, our network always takes an input image of size $640\times480$. We follow the same practice to resize larger images as~\cite{Brachmann_2018_CVPR}. That is, the image is first rescaled to height  480. 
If its width is still larger than 640, it is cropped to width 640. 
Central cropping is used at test time, and random horizontal offsets are applied during training.


\subsection{Network Training}


For 7-Scenes and 12-Scenes, our network is trained from scratch for 300K iterations with an initial learning rate of $10^{-4}$ using Adam~\cite{Adam}, and the batch size is set to 1. We halve the learning rate every 50K iterations for the last 200K iterations.  For the Cambridge Landmark dataset,  the dense reconstructions are far from perfect.  
The rendered ground truth scene coordinates contain a significant amount of outliers, which make the training difficult. 
Therefore, we train the network for 600K iterations for experiments on this dataset. For the combined scenes, the network is trained for 900K iterations.  

As mentioned in the main paper, we found that the accuracy of the final regression predictions is critical to high localization performance. Therefore, a larger weight is given to the regression loss term.
The weights for the classification loss terms $w_1$, $w_2$ are set to 1 for all scenes. The weight for the regression loss term is set to 100,000 for the three combined scenes and 10 for the other datasets.

For data augmentation,  affine transformations are applied to each training image. 
We translate, rotate, scale, shear the image by values uniformly sampled from [-20\%, 20\%], [$-30^{\circ}$, $30^{\circ}$], [0.7,1.5], [$-10^{\circ}$, $10^{\circ}$], respectively. In addition, we also augment the images with additive brightness changes uniformly sampled from [-20, 20]. When training without data augmentation, as with~\cite{Brachmann_2018_CVPR}, we randomly shift the image by -4 to 4 pixels, both horizontally and vertically, to make full use of data, as the output resolution is reduced by a factor of 8.

\subsection{Pose Optimization}

At test time, we follow the same PnP-RANSAC pipeline and  parameter settings as in~ \cite{Brachmann_2018_CVPR}. 
The inlier threshold is set to $\tau=10$ for all the scenes. 
The softness
factor is set to $\beta=0.5$  for the soft inlier count~\cite{Brachmann_2018_CVPR}.   A set of 256 initial hypotheses are sampled, and  the refinement of the selected hypothesis is performed until convergence for a maximum of 100 iterations.

\subsection{Run Time}

The network training takes $\approx$12 hours for 300K iterations on an NVIDIA Tesla V100 GPU, and $\approx$18 hours on an NVIDIA GeForce GTX 1080 Ti GPU. 

At test time, it takes $\approx$100ms for our method to localize an image on an NVIDIA GeForce GTX 1080 Ti GPU and an Intel Core i7-7820X CPU. 
Scene coordinate prediction takes 50-65ms depending on the network size. Pose optimization takes 30-60ms depending on the accuracy of the predicted correspondences.


\section{Experiments on the Aachen Dataset}

In this section, we provide the experimental details on the Aachen dataset.
\subsection{Ground Truth Labels}

Similar to the experiments on the other datasets, to generate the ground truth location labels, we run hierarchical k-means clustering on the sparse point cloud model used in~\cite{sarlin2019coarse}, which is built with COLMAP~\cite{schoenberger2016sfm,schoenberger2016mvs} using SuperPoint~\cite{DeTone_2018_CVPR_Workshops} as local feature detector and descriptor.
For this dataset we adopt a  4-level classification-only network. We also experimented with two classification-regression networks, but the  4-level classification-only network works better (see \tab{aachen} in the main paper).
For the 4-level classification-only network, we use four-level hierarchical k-means with the branching factor set to 100 for all levels. 
This results in $\approx$685K valid clusters at the finest level, with each of them containing only a single 3D point. For the experiments with the 4-level classification-regression network and the 3-level classification-regression network, we use three-level and two-level hierarchical k-means with the same branching factor setting (100 for all levels), respectively.
\subsection{Network Architecture}

\begin{figure*}[ht]
  \includegraphics[width=\linewidth]{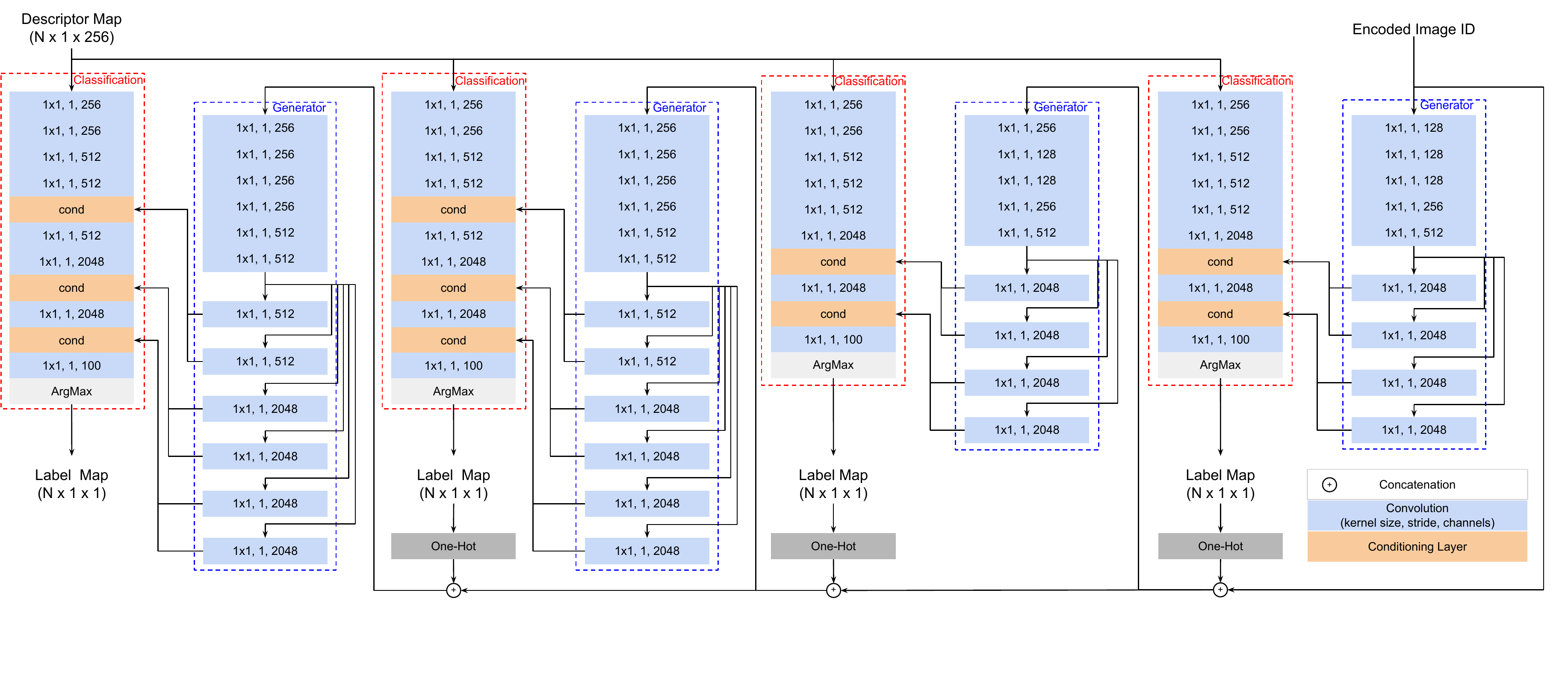}
   \centering
  \caption{Detailed network architecture of the 4-level classification-only network for the Aachen dataset experiments.}
  \label{fig:net3}
\end{figure*}

As stated in the main paper, for the experiments on the Aachen dataset, we use a list of sparse features as input to the network, rather than a regular RGB image.
Due to the sparse and irregular format of the input, we use  $1\!\times\!1$  convolutional layers in the network. We add a dummy spatial dimension to the input, \ie we use a descriptor map of size N$\times$1$\times$256 as input. In addition, there are no shared layers between different levels. To use image-level contextual information, every output layer including the first one is also conditioned on an image ID. 
To achieve this, the encoded image ID is concatenated with the label maps (if available) and then fed into the conditioning parameter generators. As mentioned in the main paper, during training, we use the ID of the training image. At inference time, we adopt NetVLAD~\cite{NetVLAD} for global image retrieval, and we use the ID of a retrieved image.
The detailed architecture of the 4-level classification-only network is given in \fig{net3}. For the 4-level classification-regression network, we simply change the last classification layer to a regression output layer.  
For the 3-level classification-regression network, one classification level is further removed.

\subsection{Network Training}


The network is trained from scratch for 900K iterations with an initial learning rate of $10^{-4}$ using Adam~\cite{Adam}, and the batch size is set to 1, similar to the previous experiments. We halve the learning rate every 50K iterations for the last 200K iterations. %
As in~\cite{Brachmann_2019_ICCV,sarlin2019coarse}, all images are converted to grayscale before extracting the descriptors. Random affine transformations, brightness and contrast changes are also applied to the images before the feature extraction. 
During training, we ignore the interest point detection, and a descriptor is extract from the dense descriptor map if it has an available corresponding 3D point in the spare 3D model. 
Following~\cite{sarlin2019coarse}, before extracting the NetVLAD~\cite{NetVLAD} and SuperPoint~\cite{DeTone_2018_CVPR_Workshops} features, the images are downsampled such that largest dimension is 960. At test time, for SuperPoint, Non-Maximum Suppression (NMS)  with radius 4 is applied to the detected keypoints and 2K of them with the highest keypoint scores are used as the input to our network.

\subsection{Pose Optimization}

We follow the PnP-RANSAC algorithm as in~ \cite{sarlin2019coarse} and the same parameter settings are used. The inlier threshold is set to $\tau=10$, and at most 5,000 hypotheses are sampled if no hypotheses with more than 100 inliers are found. Note that the pose optimization is applied independently for all the top-10 retrieved database images.

\subsection{Run Time}

The network training takes 2-3 days on an NVIDIA Tesla V100/NVIDIA GeForce GTX 1080 Ti GPU. 
On an NVIDIA GeForce GTX 1080 Ti GPU and an Intel Core i7-7820X CPU, it takes $\approx$1.1/1.4s (Aachen Day/Aachen Night) for our method to localize an image. 
It takes $\approx$170ms to extract the global and local descriptors. 
Scene coordinate prediction takes $\approx$280ms (10$\times$28ms)  and pose optimization takes $\approx$600/900ms (10$\times$60/90ms) (Aachen Day/Aachen Night). 
The time needed for global descriptor matching and the simple pre-RANSAC filtering is negligible. 

\section{Additional Qualitative Results}
\begin{figure*}[ht]
  \includegraphics[width=\linewidth]{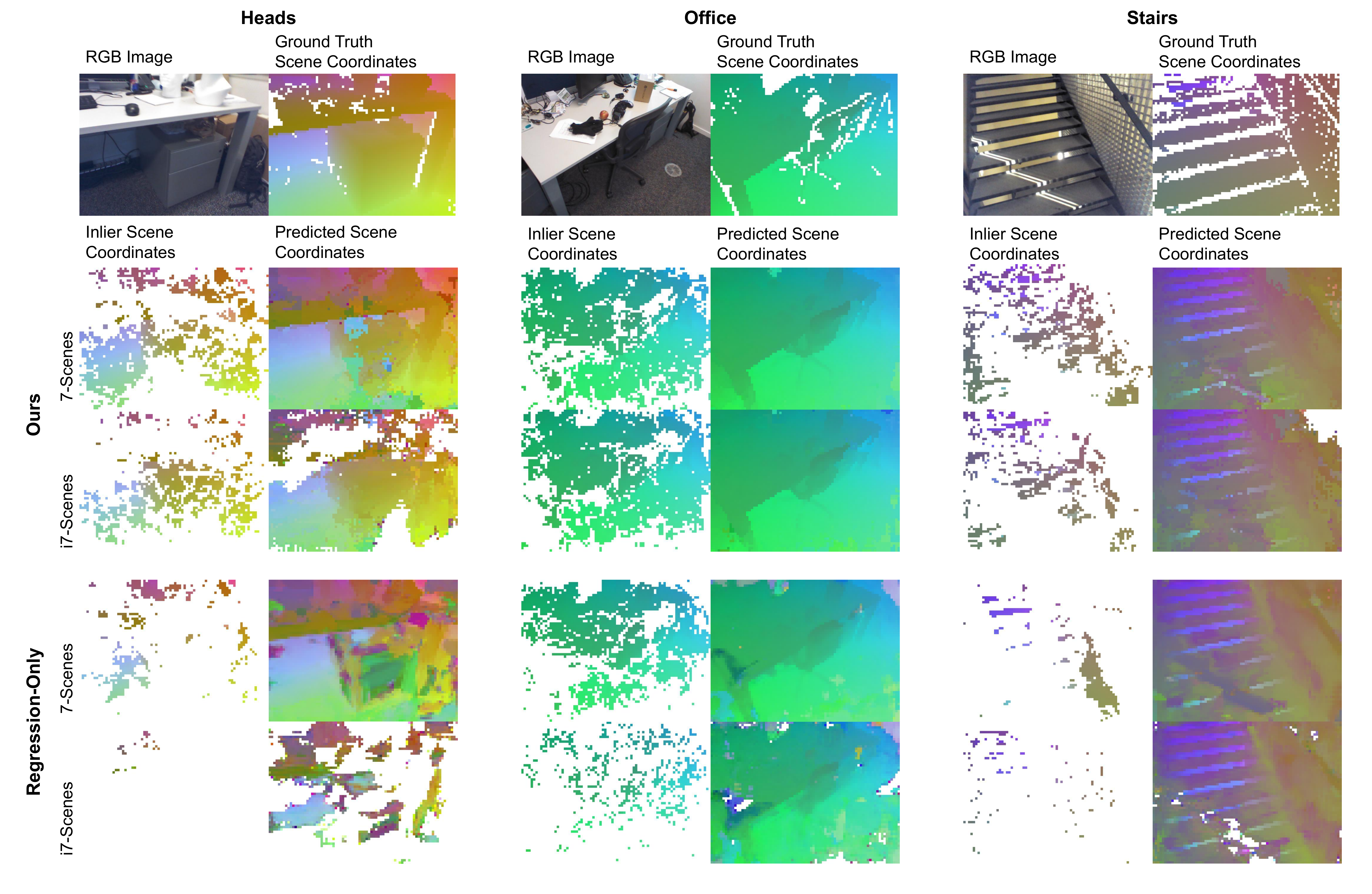}
   \centering
  \caption{We visualize the scene coordinate predictions for three test images from 7-Scenes/i7-Scenes. The XYZ coordinates are mapped to RGB values. The ground truth scene coordinates are computed from the depth maps, and invalid depth values (0, 65535) are ignored. Should a scene coordinate prediction be out of the scope of the corresponding individual scene,  the prediction is treated as invalid and not visualized. We also visualize the inlier scene coordinates retained after the pose optimization (PnP-RANSAC) stage. On both 7-Scenes and i7-Scenes, our method produces consistently better scene coordinate predictions with more inliers compared to the regression-only baseline.}
  \label{fig:q1}
\end{figure*}

We show in \fig{q1}  the quality of scene coordinate predictions for test images from 7-Scenes/i7-Scenes, and compare our method to the regression-only baseline. The scene coordinates are mapped to RGB values for  visualization. 

\begin{figure*}[ht]
  \includegraphics[width=\linewidth]{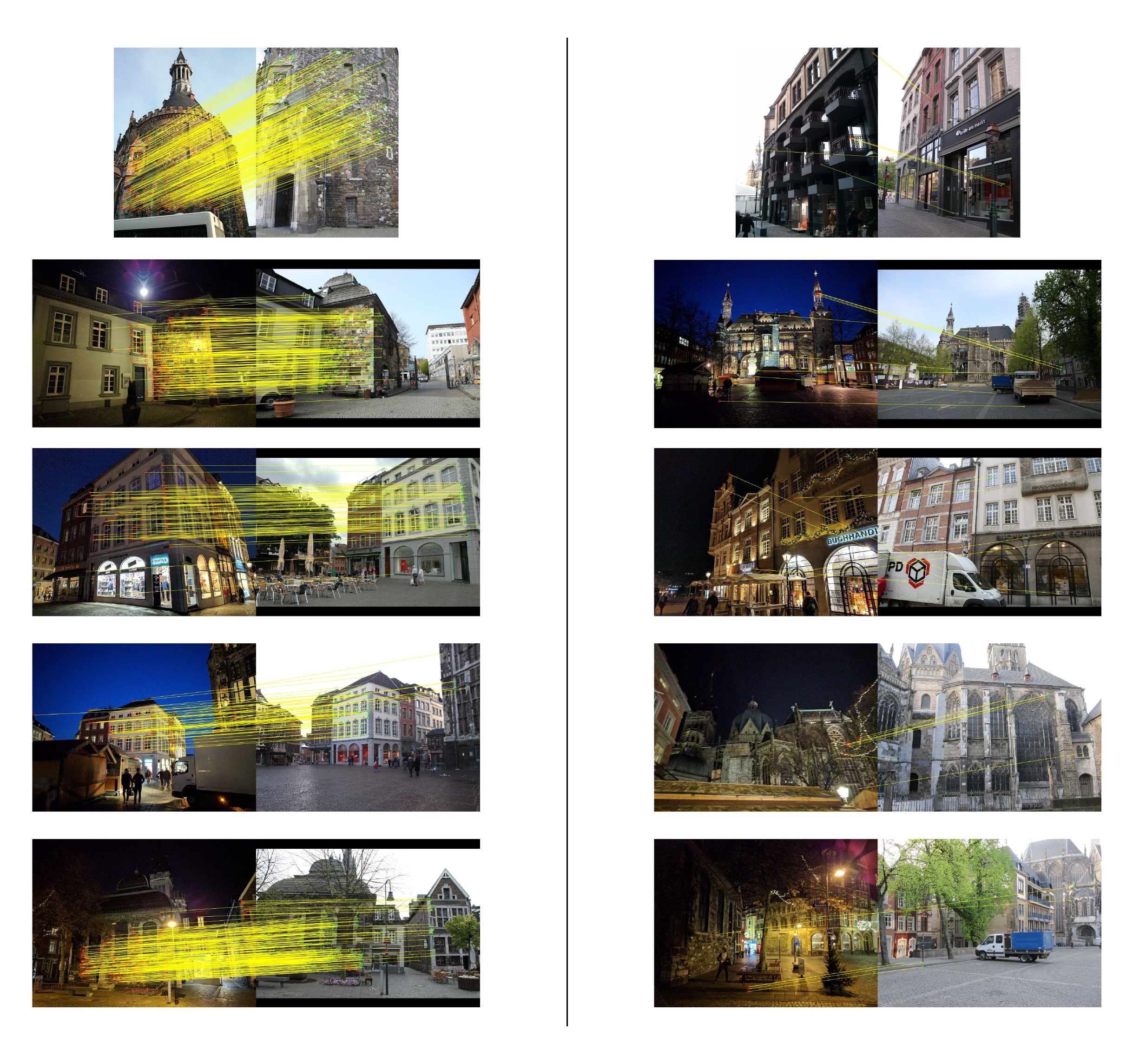}
   \centering
  \caption{We show the scene coordinate predictions for the Aachen dataset experiments. The scene coordinate predictions are visualized as 2D-2D matches between the query (left) and database (right) images. For each pair, the retrieved database image with the largest number of inliers is selected, and only the inlier matches are visualized.  We show that our method is able to produce accurate correspondences for challenging queries (left column). Failure cases are also given (right column).}
  \label{fig:q2}
\end{figure*}

We show in \fig{q2} the quality of scene coordinate predictions for the Aachen dataset experiments. The scene coordinate predictions are visualized as 2D-2D matches between the query and database images. We show only the inlier matches. 

\end{document}